\documentclass[10pt, a4paper]{article}

\usepackage{lrec-coling2024} 
\usepackage{multirow}

\title{Validating and Exploring Large Geographic Corpora}

\name{Jonathan Dunn} 

\address{University of Illinois Urbana-Champaign \\
         Department of Linguistics \\
         jedunn@illinois.edu\\
         }

\abstract{
This paper investigates the impact of corpus creation decisions on large multi-lingual geographic web corpora. Beginning with a 427 billion word corpus derived from the Common Crawl, three methods are used to improve the quality of sub-corpora representing specific language-country pairs like New Zealand English: (i) the agreement of independent language identification systems, (ii) hash-based deduplication, and (iii) location-specific outlier detection. The impact of each of these steps is then evaluated at the language level and the country level by using corpus similarity measures to compare each resulting corpus with baseline data sets. The goal is to understand the impact of upstream data cleaning decisions on downstream corpora with a specific focus on under-represented languages and populations. The evaluation shows that the validity of sub-corpora is improved with each stage of cleaning but that this improvement is unevenly distributed across languages and populations. This result shows how standard corpus creation techniques can accidentally exclude under-represented populations.\\ \newline \Keywords{geographic corpora, web corpora, outlier detection, deduplication, language identification} }

\begin{document}

\maketitleabstract

\section{Multi-Lingual Geographic Corpora}

This paper investigates the impact of corpus creation methods on large multi-lingual geographic corpora. A \textit{geographic} corpus here is a collection of written language which represents the production of a particular place, defined at the country level. Thus, these geographic corpora might distinguish between Australian English and Indian English using a combination of language identification (to determine that each sample represents English) and geo-referencing (to determine that a sample represents Australia or India). The \textit{place} which a corpus represents is really a proxy for the \textit{population} which that corpus represents. In other words, a corpus from Australia would capture the production of people currently in Australia, whether long-term locals or recent immigrants. Such multi-lingual geographic corpora are important for NLP if the field endeavours to equally represent all languages and populations. A list of the largest non-English language-country sub-corpora in the final data set is shown in Table \ref{tab:pairs} at the end of this paper.\footnote{These corpora are visualized and available for download at \href{https://www.earthLings.io}{https://www.earthLings.io}}

This paper starts with an existing web corpus derived from the Common Crawl, the 427 billion word \textit{Corpus of Global Language Use} (\textsc{cglu}: \citealt{Dunn2020}). The paper then implements a series of three improved data cleaning methods: \textit{First}, the use of multiple language identification models to triangulate the actual language of ambiguous texts; \textit{Second}, the use of hash-based deduplication to further remove samples dominated by non-authentic data (i.e., boilerplate texts which do not represent the linguistic production of individuals); \textit{Third}, perplexity-based outlier detection which is specific to a text's place of origin (i.e., so that Swiss German is not seen as an outlier of other dialects of German). 

Each of these improved cleaning methods is then evaluated across languages and across countries in order to understand both their wider impact on the corpus as a whole but also the distribution of their impact across different populations. In other words, a cleaning method which improves a corpus of Swiss German might at the same time degrade a corpus of Hindi from Fiji. The goal is to localize the impact of each of these methods. This paper is the first systematic investigation of how computational corpus creation methods influence the amount of linguistic diversity contained in large corpora.\footnote{The full supplementary material is available at \href{https://doi.org/10.17605/OSF.IO/A26RH}{DOI: 10.17605/OSF.IO/A26RH}}

\begin{figure*}[t]
\centering
\includegraphics[width = 450pt]{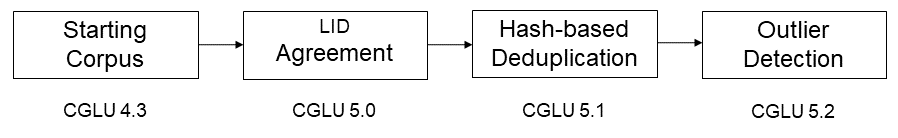}
\caption{Sequence of cleaning methods from \textsc{cglu} 4.3 to \textsc{cglu} 5.2}
\label{fig:workflow}
\end{figure*}

The basic idea behind the evaluation method is to compare the resulting geographic corpora (i.e., Swiss German on the web) to two types of baseline corpora: First, ground-truth corpora used to train language identification models; Second, geo-located tweets representing the same local populations (such as tweets in German from Zurich). The first ground-truth corpus evaluates the \textit{language} the corpus represents and the second evaluates the \textit{population} the corpus represents. Our hypothesis would be that each additional cleaning method should increase the similarity between the web corpus and these baseline corpora. One contribution of this work is the systematic use of baseline corpora of known quality to validate larger geographic corpora while not discarding samples from under-represented populations. The challenge is to maintain linguistic diversity while still removing noise.

\begin{table*}[t]
\centering
\begin{tabular}{|l|r|r|r|r|r|}
\hline
\textbf{Region} & \textbf{\textsc{cglu} 4.3} & \textbf{\textsc{cglu} 5.0} & \textbf{\textsc{cglu} 5.1} & \textbf{\textsc{cglu} 5.2} & \textbf{\textsc{twglu}} \\
\hline
Africa, North & 1,245 million & 1,064 million & 666 million & 646 million & 466 million \\
Africa, Southern & 42 million & 38 million & 16 million & 15 million & 382 million \\
Africa, Sub & 6,23 million & 5,660 million & 4,267 million & 4,241 million & 1,349 million \\
\hline
America, Brazil & 2,265 million & 1,955 million & 751 million & 718 million & 492 million \\
America, Central & 8,972 million & 6,111 million & 3,339 million & 3,259 million & 2,311 million \\
America, North & 51,921 million & 36,035 million & 26,964 million & 26,391 million & 1,155 million \\
America, South & 22,441 million & 17,678 million & 11,720 million & 11,459 million & 2,791 million \\
\hline
Asia, Central & 7,185 million & 5,960 million & 4,155 million & 4,053 million & 473 million \\
Asia, East & 49,521 million & 37,695 million & 7,901 million & 7,389 million & 924 million \\
Asia, South & 15,147 million & 10,701 million & 7,515 million & 7,341 million & 2,738 million \\
Asia, Southeast & 23,189 million & 20,225 million & 15,118 million & 14,563 million & 1,218 million \\
\hline
Europe, East & 65,413 million & 54,544 million & 38,455 million & 37,261 million & 1,372 million \\
Europe, West & 146,327 million & 111,617 million & 80,289 million & 77,825 million & 5,547 million \\
Europe, Russia & 15,363 million & 12,561 million & 11,455 million & 11,240 million & 324 million \\
\hline
Middle East & 11,606 million & 10,378 million & 7,239 million & 7,102 million & 1,224 million \\
Oceania & 329 million & 278 million & 137 million & 134 million & 1,223 million \\
\hline
\textbf{TOTAL} & \textbf{427 billion} & \textbf{332 billion} & \textbf{219 billion} & \textbf{213 billion} & \textbf{23 billion} \\
\hline
  \end{tabular}
  \caption{Size of corpora in number of words by region. \textsc{cglu 4.3} is the starting point. \textsc{cglu 5.0} includes two language identification models. \textsc{cglu 5.1} includes hash-based deduplication at the sentence level. \textsc{cgluv 5.2} includes location-sensitive outlier detection. \textsc{twglu 3.0} is the baseline for geo-referencing.}
  \label{tab:1}
\end{table*}

The evaluation thus considers a sequence of four large geographic corpora, as shown in Table \ref{tab:1}. The starting point contains 427 billion words (\textsc{cglu} 4.3). After multiple language identification models are applied this is reduced to 332 billion words (\textsc{cglu} 5.0). After hash-based deduplication at the sentence level this is reduced to 219 billion words (\textsc{cglu} 5.1). And after location-specific outlier detection this is reduced to 213 billion words (\textsc{cglu} 5.2). The size of the twitter baseline is 23 billion words (\textsc{twglu}). While this shows the overall magnitude of each additional cleaning step, the analysis in the paper is focused on two questions: First, what is the distribution of the impact of each step across languages and across countries? Second, do these cleaning steps increase the validity of the resulting geographic corpora?

This paper makes two primary contributions: In the first case, the analysis here provides a systematic evaluation of corpus creation methods across many languages and many populations. This is important because the uneven impact of methods shown in this paper implies that existing approaches to removing noise from large corpora also systematically remove sources of linguistic diversity. In the second case, this paper describes a new publicly-available multi-lingual geographic corpus which is cleaner and more usable than the original data from the Common Crawl. The general workflow for the successive versions of the corpus evaluated here is shown in Figure \ref{fig:workflow}. This figure connects the specific cleaning steps with the relevant version of \textsc{cglu}; these versions will be used throughout the paper as a shorthand. The three main steps are (i) enforcing language agreement (\textsc{cglu} 5.0), (ii) improved deduplication (\textsc{cglu} 5.1), and (iii) improved outlier detection (\textsc{cglu} 5.2).

The paper is organized as follows: in Section 2 we consider related work on geographic corpora in particular and corpus validation more broadly. In Section 3 we consider the sources of data for the geographic web corpus as well as for the benchmark corpora used for purposes of comparison. We then describe and evaluate the impact of using independent language identification models to validate language labels (Section 4), of using hash-based deduplication (Section 5), and of using location-specific outlier detection (Section 6). Finally, in Section 7 we use the benchmark corpora to measure improvements to the corpus as a whole and in Section 8 we consider the wider implications for both multi-lingual and geographic corpora.

\section{Related Work}

There is a long history of using geographic corpora to represent language use by different populations, including country-specific corpora like the \textit{British National Corpus} \citep{Consortium2007} and the \textit{Corpus of Contemporary American English} \citep{Davies2008}. While these focus on individual countries, the \textit{International Corpus of English} focuses instead of comparing a single language across multiple populations \citep{Greenbaum1996}. While these early corpora were somewhat limited in size, more recent corpora like the \textit{Corpus of Global Web-based English} \citep{Davies2013} include web-crawled data to create a total corpus of 1.9 billion words -- somewhat small for NLP but significantly larger than previous geographic corpora. Multi-lingual geographic corpora soon followed, including the 427 billion word \textit{Corpus of Global Language Use} \citep{Dunn2020} and \textit{GeoWAC} \cite{dunn-adams-2020-geographically}, a collection of gigaword corpora for 50 languages, each balanced geographically to represent the real-world distribution of speakers for that language.

\begin{table*}[t]
\centering
\begin{tabular}{|l|r|r|r|r|r|}
\hline
\textbf{Language} & \textbf{\textsc{cglu} 4.3} & \textbf{\textsc{cglu} 5.0} & \textbf{\textsc{cglu} 5.1} & \textbf{\textsc{cglu} 5.2} & \textbf{Change} \\
\hline
Russian (rus) & 5.96\% & 6.57\% & 8.73\% & 8.80\% & +2.83\% \\
Vietnamese (vie) & 3.76\% & 4.60\% & 5.67\% & 5.59\% & +1.82\% \\
German (deu) & 4.83\% & 5.28\% & 6.53\% & 6.53\% & +1.69\% \\
French (fra) & 6.15\% & 6.38\% & 7.20\% & 7.17\% & +1.02\% \\
Dutch (nld) & 2.23\% & 2.42\% & 3.11\% & 3.14\% & +0.90\% \\
Polish (pol) & 1.51\% & 1.71\% & 2.26\% & 2.25\% & +0.73\% \\
Farsi (fas) & 2.30\% & 2.78\% & 3.00\% & 3.03\% & +0.73\% \\
\hline
Finnish (fin) & 1.02\% & 0.75\% & 0.80\% & 0.79\% & -0.22\% \\
Spanish (spa) & 9.06\% & 8.23\% & 8.61\% & 8.55\% & -0.51\% \\
Portuguese (por) & 1.45\% & 1.16\% & 0.95\% & 0.94\% & -0.51\% \\
Serbo-Croatian (hbs) & 1.81\% & 1.72\% & 1.18\% & 1.19\% & -0.62\% \\
Japanese (jpn) & 3.62\% & 3.88\% & 1.53\% & 1.46\% & -2.16\% \\
English (eng) & 30.21\% & 26.62\% & 26.46\% & 26.61\% & -3.59\% \\
Chinese (zho) & 5.74\% & 6.68\% & 1.16\% & 1.14\% & -4.59\% \\
\hline
  \end{tabular}
  \caption{Size of corpora by language, given in percent of corpus in number of words after character segmentation. \textsc{cgluv4.3} is the starting point. \textsc{cglu5.0} includes two language identification models. \textsc{cglu5.1} includes hash-based deduplication at the sentence level. \textsc{cgluv5.2} includes location-specific outlier detection. \textsc{change} is the difference between the original corpus and the final corpus.}
  \label{tab:2}
\end{table*}

Geographic corpora explicitly represent different populations of speakers and thus have often been used to study linguistic variation. Such studies also provide a method of validating the corpora themselves. For instance, previous work has shown that both geographic web corpora \citep{Cook2017} and geo-referenced tweets  \citep{10.3389/frai.2019.00011} can be used to replicate traditional studies of lexical variation across dialects. At the syntactic level, another line of work has shown that such corpora can be used to model grammatical variation across dialects \citep{d18b, dunn-2019-modeling, 10.3389/frai.2019.00015, 10.3389/fcpxs.2023.1273741}. Such work provides a validation of geographic corpora, at least in major languages like English and French, in the sense that they contain the linguistic variants expected from dialect surveys.

Another approach to validating geographic corpora relies on corpus similarity methods \citep{Kilgarriff2001, fothergill-etal-2016-evaluating} to compare geo-referenced corpora to known ground-truth data from alternate sources. Recent work has shown that such measures can be updated for highly multi-lingual corpora \citep{Li2022a} and also used to differentiate corpora from different sources or registers \citep{Li2022}. Recent work \citep{Dunn2021} has applied these methods to geographic corpora by using geo-referenced tweets (with quite precise geographic meta-data) to validate geo-referenced web pages (which rely on less reliable top-level domain information). The question, again, is whether a geographic corpus actually represents the underlying population it is meant to represent. Another recent approach to exploring geo-referenced corpora relies on comparing named entities extracted from the text with the expected location of such entities \citep{faisal-etal-2022-dataset}. While promising, such an approach relies on both the extraction of named entities and a reliable gazetteer of their location, both of which are problematic for low-resource languages and under-represented locations.

Recent audits of multi-lingual corpora have shown that many automated efforts create problematic corpora \citep{kreutzer-etal-2022-quality}. In part this is because of upstream issues in the data used for language identification models; given the enormity of digital sources like the web, language identification models with a relatively high precision and recall on formal texts can produce corpora which do not actually represent that language when applied to informal texts. This is one motivation for applying multiple language identification models to provide increased certainty about language labels.

This current work can also be situated relative to other corpus creation pipelines that start with the Common Crawl, such as CCNet \citep{wenzek-etal-2020-ccnet}. An important rationale is to measure the impact of each design decision, not only in the aggregate but also across languages and across populations. For instance, the outlier detection employed in CCNet relies on language models trained on Wikipedia, even though we know that there is a strong geographic skew in the contributors to Wikipedia \citep{Graham2015} and even though there is a significant linguistic difference between Wikipedia's register and others \citep{Li2022}.

There is a trade-off here between (i) producing clean and usable corpora which contain utterances that are meaningful to human speakers while (ii) capturing linguistic variation and allowing for non-standard forms from under-represented populations. For instance, it has been shown that a large English corpus from the Common Crawl excludes certain populations as a result of cleaning decisions \citep{dodge-etal-2021-documenting}. The goal of this paper, then, is to measure the impact of each stage of processing across languages and populations.

\section{Data and Methods}

The main data which serves as the starting point for this analysis is the full \textit{Corpus of Global Language Use} (\textsc{cglu}; \citealt{Dunn2020}). This is a 427 billion word web corpus derived from the Common Crawl and geo-referenced mainly using information from top-level domains. The corpus is organized into 16 larger regions as shown in Table \ref{tab:1}; the same geographic structure is maintained in this paper in order to enable the analysis of the impact of different cleaning steps.

Two sets of comparison corpora are used for validation. First, ground-truth language identification corpora are taken from seven independent sources: Bible translations \cite{Brown2014}, Global Voices News \cite{Tiedemann2012}, the JW 300 data set \cite{agic-vulic-2019-jw300}, Open Subtitles \cite{lison-tiedemann-2016-opensubtitles2016}, QCRI Educational Domain \cite{Tiedemann2012},  Tatoeba Sentences \cite{Tiedemann2012}, and Wikipedia articles. Samples are drawn equally from these sources to avoid over-representing a specific register. Known inconsistencies for several languages in the JW 300 data are corrected \citep{kreutzer-etal-2022-quality}. These samples represent corpora from a known language. The second source of comparison corpora is tweets collected from the same countries as the original \textsc{cglu} and divided into regions in the same manner \citep{10.3389/frai.2019.00015, dunn-wong-2022-stability, Dunn2023bb}. The distribution of this corpora, totalling 23 billion words, is shown in Table \ref{tab:1}. Because language identification is an essential component of preparing these types of corpora, the same models are used as for the agreement stage in Section 4. These samples represent corpora from a known location.

To compare each stage of cleaning with these benchmark corpora we use existing corpus similarity measures \citep{Li2022, Li2022a}. These measures rely on the frequency of character n-grams to differentiate between more or less similar corpora. Previous work has evaluated these in a cross-register and multi-lingual setting; here we rely directly on the models provided from this previous work.\footnote{\href{https://www.github.com/jonathandunn/corpus_similarity}{https://github.com/jonathandunn/corpus\_similarity}} These comparisons are limited to the 74 languages covered by the existing package.

\begin{figure*}[t]
\centering
\includegraphics[width = 475pt]{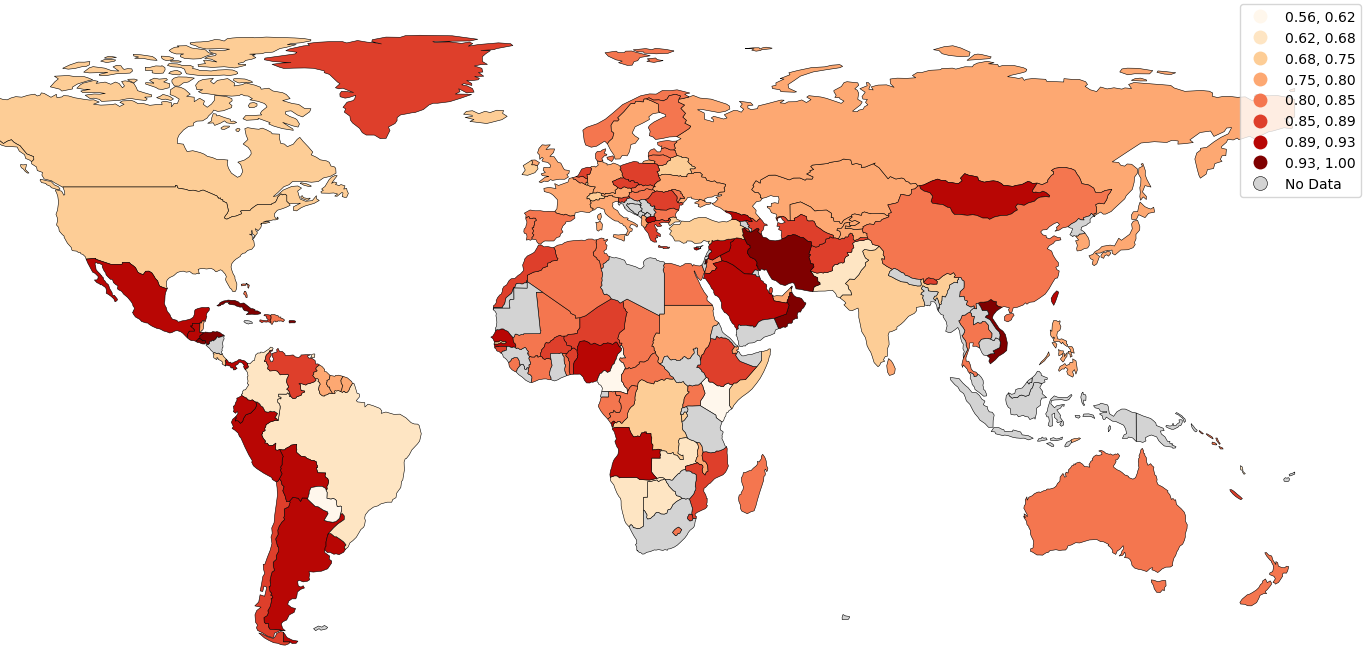}
\caption{Map showing agreement between language identification models by country. A value of 0.80 means that 80\% of samples receive the same language label from each model.}
\label{fig:map}
\end{figure*}

\section{Validating Language Labels}

Classifying documents by language is an essential part of creating a web corpus. And yet the problem of language identification itself remains a challenge for less-common languages \citep{Brown2014}, for non-standard varieties of languages \citep{jurgens-etal-2017-incorporating}, and for less-common or informal registers \citep{lui-baldwin-2011-cross}. Less-common languages often have lower precision or recall \citep{jauhiainen-etal-2022-heli} and this can have downstream ramifications for corpus quality \citep{kreutzer-etal-2022-quality}. Our first validation step, then, is to use an independent language identification model to label the original \textsc{cglu} data. The original paper used a feed-forward network for classification \citep{Dunn2020}. We employ a more recent fastText-based model \citep{dunn-nijhof-2022-language}. Importantly, the second model represents all 464 languages that are represented by the first model.

Given two language labels for each document (from the original model and the more recent model), cases of agreement provide greater confidence that the language has been correctly identified. Noisy samples, especially, will be miscategorized by one or both models but are less likely to be miscategorized into the same label. The distribution of agreement rates across countries is shown in Figure \ref{fig:map}, with darker red indicating high agreement and lighter red low agreement. We see from the map that the impact of this language validation step is not distributed equally: North America and Brazil, for instance, have much lower agreement rates than countries in Central and South America.

An alternate view is provided in Table \ref{tab:2} which shows the percentage of the web corpus which is composed of specific languages through each step. The difference between \textsc{cglu} 4.3 and \textsc{cglu} 5.0, then, reflects the constraint that all documents whose language labels disagree are removed. Some languages, like Russian and Vietnamese, increase in their relative share of the data. Others, however, decrease: especially very common languages like English and Spanish.

\begin{figure*}[t]
\centering
\includegraphics[width = 450pt]{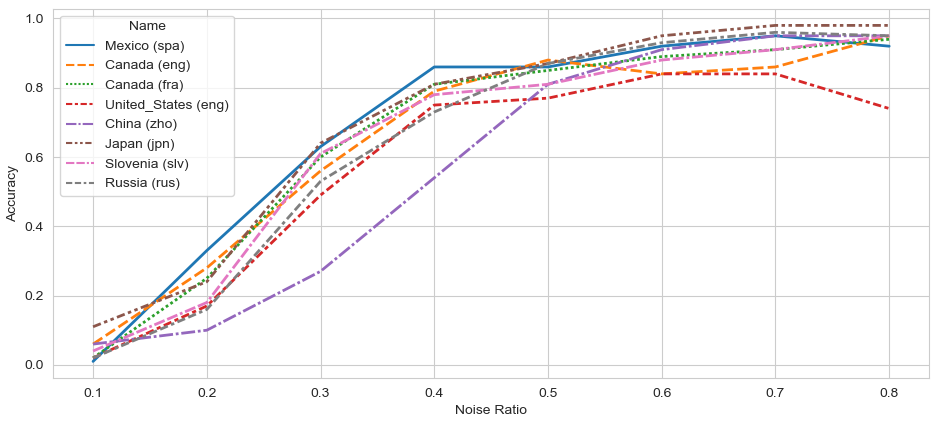}
\caption{Accuracy of the outlier detection method for finding samples with injected noise. \textit{Ratio} refers to the amount of noise added and \textit{Accuracy} to the percent of such samples correctly identified.}
\label{fig:noise}
\end{figure*}

A decrease in majority languages or in Western countries is acceptable insofar as these languages and populations are already over-represented in the corpus. As shown in Table \ref{tab:1}, North America and Western Europe remain the single largest sources in the final corpus at 26.3 billion and 77.8 billion words, respectively. Enforcing agreement in language labels across models removes 95 billion words, or 22.2\% of the original data.

\section{Hash-based Deduplication}

The second stage of cleaning investigated here is the removal of similar samples. The original \textsc{cglu} corpus used a method based on exact matches at the level of paragraph tags, a conservative approach which avoids removing non-duplicates but has the potential to leave a large number of highly similar documents. Here we implement a more recent hash-based strategy \citep{wenzek-etal-2020-ccnet}, applied so that both line breaks and paragraph tags can separate samples. This approach first hashes each sample using the SHA-1 hashing function and then completely removes any samples which collide within that hashing space. As shown in Table \ref{tab:1}, deduplication removes 113 billion words, or 34\% of the data after language validation.

This hash-based depulication step has the biggest overall impact on the distribution of the corpus across languages and countries. We can measure this by viewing the corpus as a collection of sub-corpora representing specific language-country pairs (like Australian English). The Pearson correlation between the size of these sub-corpora after validating the language labels is 0.99 and highly significant. The correlation after deduplication is 0.93, still highly significant. And the correlation after outlier detection (see below) is again 0.99 and highly significant. Thus, deduplication is the stage which most alters the distribution of the corpus across languages and countries.

\begin{table*}[t]
\centering
\begin{tabular}{|l|l|}
\hline
Low & bed princess canopy new baby bed mosquito net cute baby princess canopy crib netting \\
Low & monster legends hack generator – monster legends hack using cheat engine monster \\
\hline\hline
High & Table 10: Brand Fortified Wine market, Table 11: Brand Sparkling Wine market \\
High & Carnival Cruise Line Celebrity Cruises Costa Cruises Disney Cruise Cruises France \\
\hline
  \end{tabular}
  \caption{Examples of outliers from the corpus of American English. \textit{High} example are too predictable according to the model. \textit{Low} examples are too unpredictable according to the model.}
  \label{tab:examples}
\end{table*}

We can see why in reference to the relative percent of each language shown in Table \ref{tab:2}: deduplication (the difference between 5.0 and 5.1) reduces Chinese from 6.68\% to 1.16\% and Japanese from 3.88\% to 1.53\%. English, which was disproportionately affected by the language validation step, has very little change during deduplication, going from 26.62\% to 26.46\%. If this were constrained to the usage of Chinese or Japanese in one or two countries, we might suspect that the data from those sources was problematic and unusually full of duplicates. The lower correlation, though, means that this pattern is spread across many individual countries. Thus, this indicates that the hash-based approach does not work equally well across languages and has the implicit side-effect of reducing some languages more than others. Future work is needed to investigate more equitable cross-linguistic deduplication methods.

\section{Outlier Detection}

The final cleaning method evaluated is outlier detection to remove noisy samples which are non-duplicate instances of the correct language. We again adapt and evaluate a recent approach that first trains a statistical language model for each language and then uses perplexity of documents to find outliers \citep{wenzek-etal-2020-ccnet}. The original approach trains a single model for each language using Wikipedia. Given that one goal of a geographic corpus is to include under-represented populations and low-resource varieties, using Wikipedia as the benchmark is likely to exclude many relevant documents. We therefore adapt this approach by instead training a language model for each language-country pair (like Swiss German) which contains at least 5 million words.\footnote{\href{https://github.com/jonathandunn/common_crawl_corpus}{github.com/jonathandunn/common\_crawl\_corpus}}

The challenge, then, is that this creates a much larger number of models to be used for outlier detection. Instead of a statistical language model with a vocabulary from a sentence piece tokenizer, we use character-based skip-gram models with a fastText base. Specifically, we use a window size of 5 with character-ngrams of length 3 to 6, trained for two epochs on each language-country sub-corpora. This model is then used to measure the log probability of documents in the corpus given the learned embeddings, the specific measure used for outlier detection. This approach is more feasible for the large number of language-country sub-corpora that need to be dealt with. Such an approach has previously been used for other document classification tasks \citep{taddy-2015-document}.

As in the original approach, then, each sample in the corpus is given a log probability so that the corpus itself is represented as a distribution of such values. We standardize these values using the Iglewicz and Hoaglin method of modifying the z-score to be robust to outliers shown in (1). This is comparable to the z-score but using the median rather than the mean and dividing by the median absolute deviation (MAD). Any sample with a standardized log probability which falls above 3 or below -3 is considered an outlier.\footnote{This method assumes a normal distribution of log probabilities within each corpus; figures in the supplementary material show that this assumption is met.}

\begin{figure*}[t]
\centering
\includegraphics[width = 450pt]{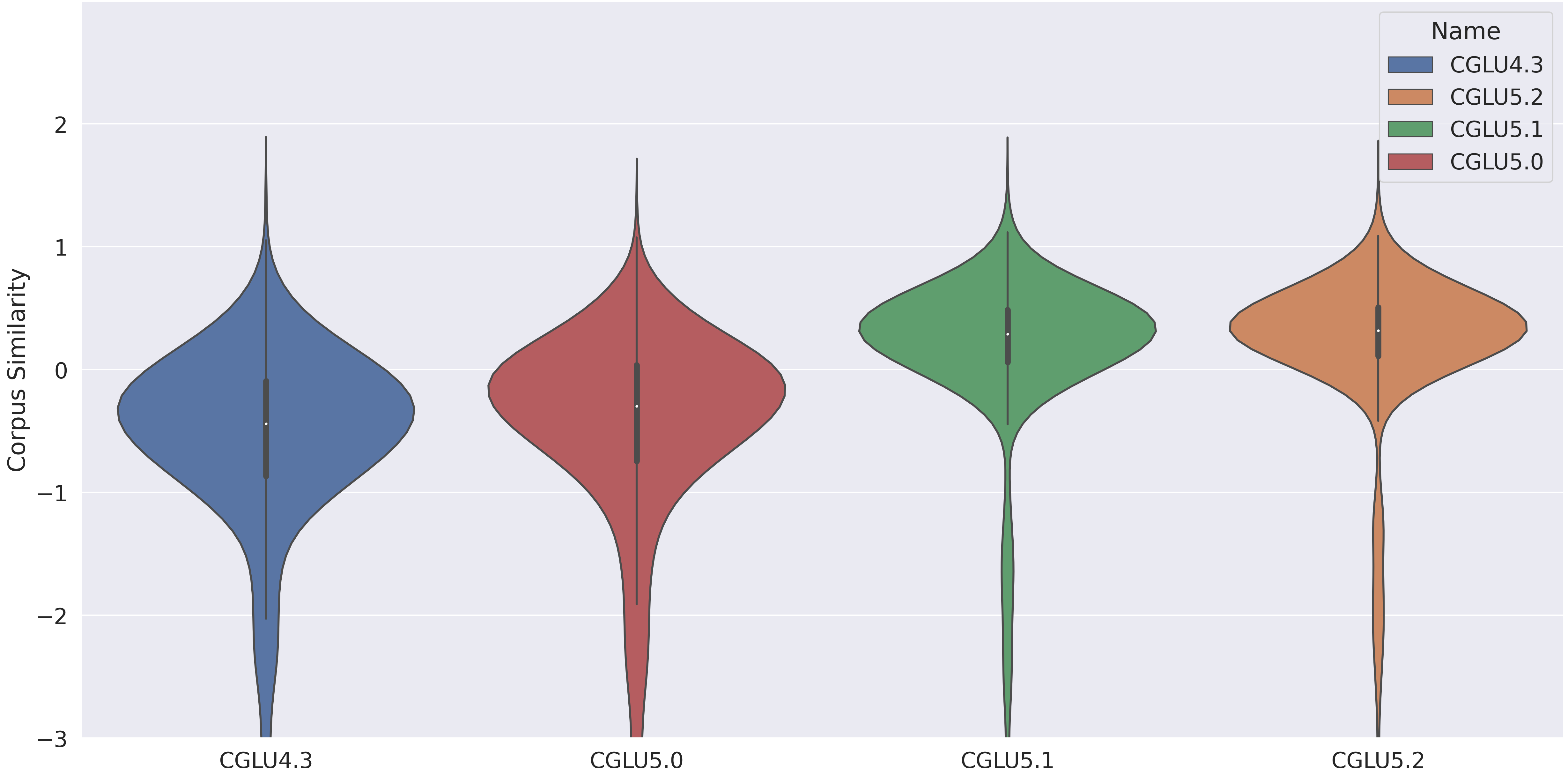}
\caption{Similarity of the Swiss German corpus to the benchmark language identification corpus over each stage of cleaning. Higher values indicate more similar corpora. Significance of differences is tested using an ANOVA, here with a value of $p<0.001$.}
\label{fig:german1}
\end{figure*}

\begin{equation}
    M_i = \frac{0.6745(x_i - median(x))}{MAD}
\end{equation}

An approach like this distinguishes between two types of outliers: those which are too unpredictable and those which are too predictable. Examples of both types are shown in Table \ref{tab:examples}. We evaluate the quality of outier detection by artificially injecting noise into samples in the form of arbitrary words from other languages. The algorithm is them evaluated on its ability to accurately identify these noisy samples. Results for eight common language-country sub-corpora are shown in Figure \ref{fig:noise}. The y-axis represents prediction accuracy, with lines near the top being able to correctly identify most or all noisy samples. The x-axis represents the amount of noise inserted into the sample. At small amounts (similar to code-switching) the noisy samples are not discarded. As the ratio increases, however, almost all noisy samples are correctly identified. This evaluation shows, then, that location-specific outlier detection is able to accurately remove noisy samples from geographic corpora.

This stage of cleaning has the smallest impact on the corpus as a whole, as shown in Table \ref{tab:1}: only 6.35 billion words are removed, or approximately 2.9\% of the data. The correlation between the distribution of the corpus before and after outlier detection is 0.999 and highly significant. Thus, this cleaning method has the smallest impact overall and is the most evenly distributed across languages and countries. This is because the model used for identifying outliers is specific to each language-country sub-corpus.

\section{Evaluation}

We turn now to the evaluation of this series of cleaning methods, from validating the language labels to hash-based deduplication to outlier detection. The basic idea in the evaluation is to take more reliable benchmark corpora which are known to represent either (i) the language in question or (ii) the population in question. We then use corpus similarity measures \citep{Li2022a} to determine if each cleaning step makes the language-country sub-corpora more similar to these benchmarks. Being more similar to the language identification corpus would mean being cleaner and more valid samples of that language. Being more similar to the geo-referenced corpus would mean being cleaner and more valid samples of that particular location. We confine this evaluation to the 74 languages covered by these measures.

We start with the comparison with benchmark language identification data, visualizing the progression of Swiss German in Figure \ref{fig:german1}. The corpus similarity measure works by breaking the corpus into 500 equal-sized chunks (each containing 10,000 words). The violin plot is thus showing the distribution of values across many pairs of 500 chunks from the web corpus and the benchmark corpus. We use an ANOVA to test whether these are in fact distinct populations, here with a significance of $p<0.001$. The violin plot shows two important features: first, the mean similarity of the corpus is increasing with each stage of cleaning; second, the tail of outliers decreases visually with each stage. This figure shows, then, that the corpus of Swiss German becomes more like expected German corpora at each stage of cleaning.

\begin{figure*}[t]
\centering
\includegraphics[width = 450pt]{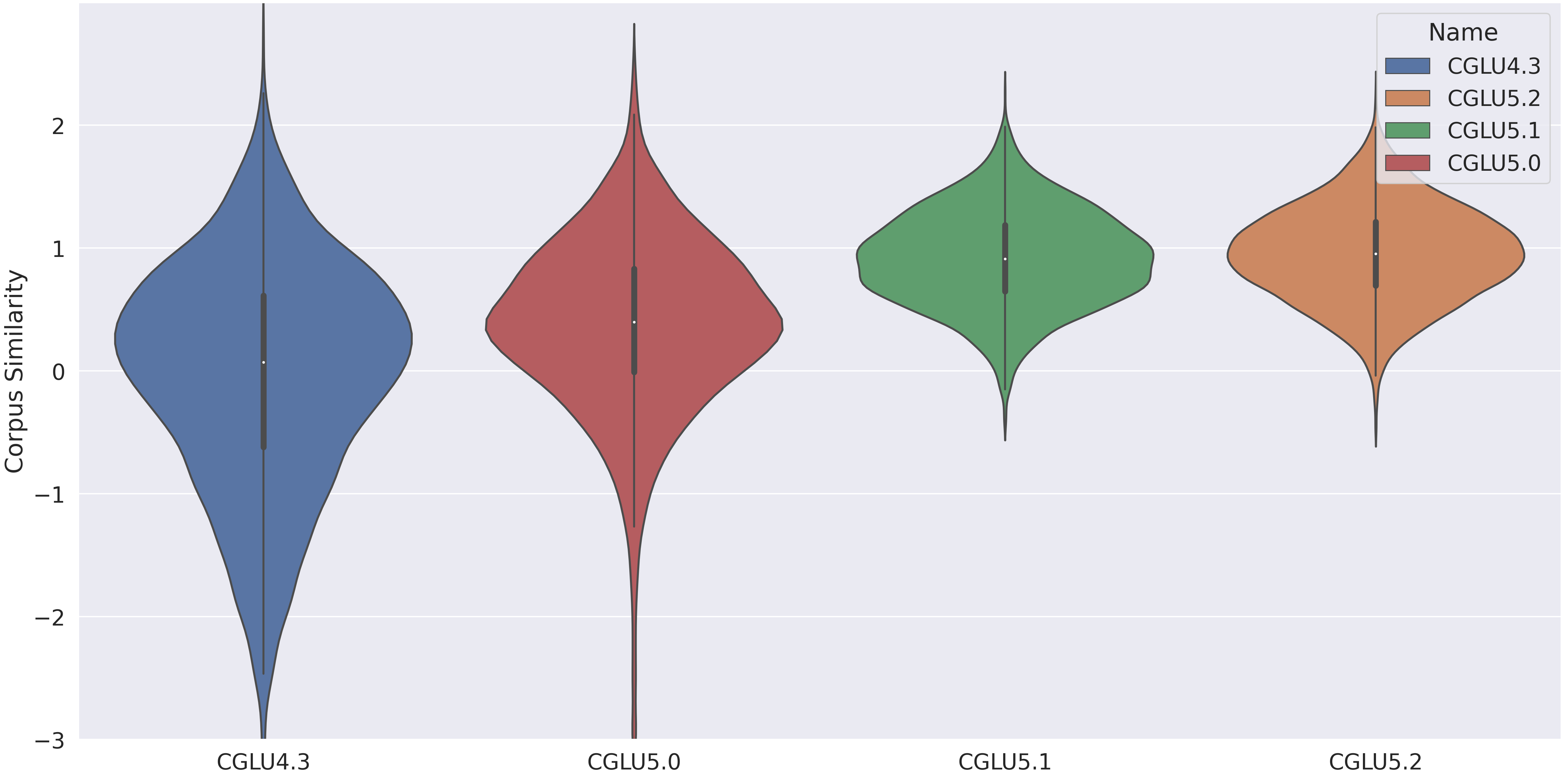}
\caption{Similarity of the Chilean Spanish corpus to the benchmark corpus of tweets in Spanish from Chile over each stage of cleaning. Higher values indicate more similar corpora. Significance of differences is tested using an ANOVA, here with a value of $p<0.001$.}
\label{fig:spanish2}
\end{figure*}

The next step compares the corpus instead with benchmark tweets. Here, in Figure \ref{fig:spanish2} we use tweets in Spanish from Chile, evaluating whether this web corpus represents the Chilean Spanish dialect rather than other varieties of Spanish. Here we see a similar progression: the mean similarity is higher with each stage of cleaning and the presence of outliers is greatly reduced. As before, the difference is significant using an ANOVA test. The deduplication stage has the largest impact, as we would expect from the fact that this stage made the largest alteration to the corpora. The outlier detection stage, with the smallest amount of removed data, has a still visible but less impactful influence on the corpus similarity measures.

The full figures across all language-country sub-corpora are available in the supplementary material and online at the \href{https://www.earthlings.io}{earthLings.io project}. In most cases there is a similar improvement as shown in Swiss German in terms of both a higher mean and a more homogenous distribution, indicating a cleaner and more valid corpus. But this is not universally true. For example, Arabic shows an improvement in geo-referencing (against tweets) across countries in both attributes. But against the language identification benchmark Arabic does not improve in the mean value and only slightly in the reduced tail of outliers. English shows improvements across the board in Canada but somewhat less consistently in places like Mexico. 

French, in contrast, shows significant improvements in geo-referencing but at the same time a slight decrease against language identification data when outside France. This implies that the language identification corpus does not adequately capture dialects of French so that accurate representations of Senegal French are actually less similar to European French. Japenese shows slight improvements in geo-referencing but no change against the language identification corpora; Chinese shows a similar pattern. Korean, in contrast, shows a significant improvement in both cases, as do most other languages. These full figures are available in the supplementary material and are useful for exploring the full impact of these cleaning methods across different languages and populations. While almost all sub-corpora improve, the degree and character of improvement varies by language and population.

\section{Discussion and Conclusions}

\begin{table*}[t]
\centering
\begin{tabular}{|lrr|lrr|lrr|}
\hline
\multirow{2}{*}{\textbf{Country}} & \multirow{2}{*}{\textbf{Lang.}} & \textbf{Words} & \multirow{2}{*}{\textbf{Country}} & \multirow{2}{*}{\textbf{Lang.}} & \textbf{Words} & \multirow{2}{*}{\textbf{Country}} & \multirow{2}{*}{\textbf{Lang.}} & \textbf{Words} \\
\textbf{~} & \textbf{~} & \textbf{(Mil)} & \textbf{~} & \textbf{~} & \textbf{(Mil)} & \textbf{~} & \textbf{~} & \textbf{(Mil)} \\
\hline
Viet Nam & vie & 11,824 & Estonia & est & 1,040 & India & rus & 121 \\
France & fra & 11,234 & Serbia & hbs & 1,011 & Albania & sqi & 117 \\
Spain & spa & 10,700 & Slovenia & slv & 1,001 & Uzbekistan & uzb & 116 \\
Russia & rus & 10,189 & Peru & spa & 769 & Luxembourg & fra & 115 \\
Germany & deu & 8,368 & Ukraine & ukr & 648 & Honduras & spa & 115 \\
Italy & ita & 7,180 & Switzerland & fra & 625 & Colombia & rus & 111 \\
Iran & fas & 6,327 & Brazil & por & 611 & Switzerland & ita & 111 \\
Netherlands & nld & 4,864 & Colombia & spa & 570 & Palau & rus & 104 \\
Poland & pol & 4,687 & Azerbaijan & aze & 466 & Malaysia & ind & 102 \\
Romania & ron & 4,295 & Bosnia & hbs & 394 & Lithuania & rus & 100 \\
Czechia & ces & 4,032 & Cuba & spa & 378 & Ecuador & spa & 97 \\
Norway & nor & 3,693 & Taiwan & zho & 365 & USA & por & 97 \\
Sweden & swe & 3,588 & South Korea & kor & 363 & Colombia & deu & 95 \\
Slovakia & slk & 3,566 & Iceland & isl & 286 & Finland & swe & 79 \\
Denmark & dan & 3,092 & Moldova & ron & 270 & Colombia & ara & 78 \\
Austria & deu & 2,971 & Indonesia & ind & 263 & USA & spa & 78 \\
Japan & jpn & 2,919 & Georgia & kat & 258 & Belarus & bel & 76 \\
Chile & spa & 2,849 & Latvia & rus & 258 & USA & fra & 74 \\
Greece & ell & 2,624 & Colombia & fra & 201 & Morocco & ara & 73 \\
Hungary & hun & 2,407 & N. Macedonia & mkd & 196 & Spain & cat & 66 \\
Lithuania & lit & 2,336 & Kazakhstan & kaz & 191 & Liechtenstein & spa & 66 \\
Belarus & rus & 2,247 & Colombia & ind & 166 & Timor-Leste & spa & 66 \\
Kazakhstan & rus & 2,144 & Uzbekistan & rus & 153 & Gabon & fra & 62 \\
Bulgaria & bul & 1,933 & Azerbaijan & rus & 152 & Colombia & jpn & 61 \\
China & zho & 1,931 & Kyrgyzstan & rus & 151 & El Salvador & spa & 59 \\
Mexico & spa & 1,844 & Ecuador & rus & 150 & C. A. Rep. & fra & 58 \\
Switzerland & deu & 1,824 & Estonia & rus & 150 & Gabon & rus & 53 \\
Ukraine & rus & 1,802 & Moldova & rus & 149 & USA & ara & 53 \\
Belgium & nld & 1,764 & Slovakia & ces & 143 & USA & deu & 53 \\
Finland & fin & 1,665 & Morocco & fra & 141 & Dominican R. & spa & 52 \\
Latvia & lav & 1,120 & USA & rus & 139 & Slovakia & hun & 52 \\
Belgium & fra & 1,095 & Haiti & rus & 134 & Colombia & por & 52 \\
Portugal & por & 1,091 & Mongolia & mon & 133 & Vatican & ita & 51 \\
Croatia & hbs & 1,080 & UAE & ara & 131 & USA & ind & 51 \\
Canada & fra & 1,062 & Palestine & ara & 127 & Réunion & fra & 51 \\
\hline
  \end{tabular}
  \caption{Size of \textsc{cgluv 5.2} for the 105 non-English language-country pairs with at least 50 million words. Each pair is a sub-corpus representing the use of a specific language in a specific country. \textit{Size} is given in millions of words per corpus.}
  \label{tab:pairs}
\end{table*}

This paper has systematically evaluated the impacts of cleaning methods on a large multi-lingual geographic web corpora, looking at the impact across both countries and languages. Three cleaning methods are evaluated in a specific order of application: (i) independent validation of language labels, (ii) hash-based deduplication, and (iii) location-specific outlier detection. 

The first two cleaning methods have a very large impact on the corpus, removing 95 billion and then 113 billion words. The first stage leaves the corpus with much the same distribution across languages and countries (a Pearson correlation of 0.99), but has more impact in some areas than others. The deduplication step has the largest impact on the distribution of the corpus, with a significant reduction in East Asian languages like Chinese and Japanese. Outlier detection has the smallest impact and does not change the distribution of the corpus; yet an evaluation shows that it remains quite accurate in removing noisy samples. These findings are important because they indicate that corpus creation methods do not have a consistent impact across languages and populations.

At the same time, it is essential to evaluate the impact of these methods because excluding documents from under-represented populations or low-resource varieties serves to further exclude these populations from the language technologies which depend on these corpora \citep{dodge-etal-2021-documenting}. In other words, excluding a document because it belongs to a low-resource variety means that variety will remain under-represented. An important contribution of geographic corpora is that they are able to ensure equal representation across diverse populations and language varieties. For instance, Table \ref{tab:pairs} shows the 105 largest non-English language-country sub-corpora contained in the final dataset. This corpus improves our representation of linguistic diversity by including many populations around the world using many languages that are often under-represented.

The main contribution of this paper is not only a systematic evaluation of these methods but also a new version of the large multi-lingual geographic web corpus, now containing 213 billion words, which is measurably better than previous versions. As shown in Table \ref{tab:1}, this corpus is organized by country and divided into 16 larger regions. While more wealthy populations from North America and Western Europe provide a large portion of this data, their combined contribution is below half of the final data set. This allows for significant representation of populations from Africa (nearly 5 billion words) and Asia (over 33 billion words), a large improvement on existing corpora.

\section{Ethics Statement}

This paper uses written digital corpora to represent diverse populations of speakers around the world. While such geographic corpora are more representative of the world's population than non-geographic corpora, it remains the case the certain demographic segments within each geographic area are more likely to contribute to the corpus. Thus, this data should not be taken as an exhaustive representation of all sub-populations within each geographic location.

\section{Bibliographical References}\label{sec:reference}

\bibliographystyle{lrec-coling2024-natbib}
\bibliography{lrec-coling2024-example}

\end{document}